\documentclass[11pt]{article}
\pdfoutput=1

\usepackage[final]{acl} % [review]

\usepackage{times}
\usepackage{latexsym}

\usepackage[T1]{fontenc}
\usepackage[utf8]{inputenc}
\usepackage{microtype}

\hypersetup{breaklinks,
            pdfauthor={Anonymous},
            pdftitle={Evaluating the Faithfulness of Importance Measures in NLP by Recursively Masking Allegedly Important Tokens and Retraining},
            pdfsubject={An evaluation of the faithfulness of importance measures, also known as saliency measure. We evaluate faithfulness on attention, gradient method, and integrated gradient. The faithfulness metric is based ROAR but with improvements. We provide a scalar benchmark for faithfulness, making comparison easy.},
            pdfkeywords={Faithfulness, Benchmark, Importance Measure, Salience, ROAR, Neural Networks, Natural Language Processing, NLP, Interpretability, Post-hoc, Attention, Gradient, Integrated Gradient}}

\usepackage{mathtools,amssymb}
\usepackage{xcolor,xstring}
\usepackage{float,subcaption,stfloats}
\usepackage{booktabs,multirow}
\usepackage{graphicx,tikz}
\usepackage{enumitem,cleveref}
\usepackage{layouts}

\usepackage{todonotes}

\renewcommand{\cite}[1]{\textcolor{red}{WRONG CITE COMMAND [\StrSubstitute{\detokenize{#1}}{,}{, }]}}

\newcommand{\importance}[1]{#1}
\newcommand{\ig}[1]{\emph{#1}}
\newcommand{\gradient}[1]{\emph{#1}}
\newcommand{\attention}[1]{\emph{#1}}

\title{Evaluating the Faithfulness of Importance Measures in NLP by Recursively Masking Allegedly Important Tokens and Retraining}

\author{Andreas Madsen\textsuperscript{1,2} \quad Nicholas Meade\textsuperscript{1,3,*} \quad Vaibhav Adlakha\textsuperscript{1,3,*} \quad Siva Reddy\textsuperscript{1,3,4} \\
  \textsuperscript{1}\,Mila -- Quebec AI Institute \quad
  \textsuperscript{2}\,Polytechnique Montréal \\
  \textsuperscript{3}\,McGill University \quad
  \textsuperscript{4}\,Facebook CIFAR AI Chair \\
  \texttt{\{firstname.lastname\}@mila.quebec} \\}

\date{}

\begin{document}
\maketitle
\makeatletter
\ifacl@finalcopy
\renewcommand*{\thefootnote}{*}
\footnotetext[0]{Equal contribution.}
\renewcommand*{\thefootnote}{\arabic{footnote}}
\setcounter{footnote}{0}
\fi
\makeatother

\begin{abstract}
To explain NLP models a popular approach is to use importance measures, such as attention, which inform input tokens are important for making a prediction. 
However, an open question is how well these explanations accurately reflect a model's logic, a property called \emph{faithfulness}.

To answer this question, we propose Recursive ROAR, a new faithfulness metric. This works by recursively masking allegedly important tokens and then retraining the model. The principle is that this should result in worse model performance compared to masking random tokens. The result is a performance curve given a masking-ratio. Furthermore, we propose a summarizing metric using relative area-between-curves (RACU), which allows for easy comparison across papers, models, and tasks.

We evaluate 4 different importance measures on 8 different datasets, using both LSTM-attention models and RoBERTa models.
We find that the faithfulness of importance measures is both model-dependent and task-dependent. This conclusion contradicts previous evaluations in both computer vision and faithfulness of attention literature.

\end{abstract}

\section{Introduction}

The ability to explain neural networks benefits both accountability and ethics when deploying models \citep{Doshi-Velez2017} and helps develop a scientific understanding of what models do \citep{Doshi-Velez2017a}. Particularly, in NLP, attention \citep{Bahdanau2015} is often used as an explanation to provide insight into the logical process of a model \citep{Belinkov2019}.

\attention{Attention}, among other methods such as \gradient{gradient} \citep{Baehrens2010,Li2016} and \ig{integrated gradient} \citep{Sundararajan2017a,Mudrakarta2018}, explain which input tokens are relevant for a given prediction. This type of explanation is called an \importance{importance measure}.

A major challenge in the field of interpretability is ensuring that an explanation is \textit{faithful}: ``a faithful interpretation is one that accurately represents the reasoning process behind the model’s prediction'' \citep{Jacovi2020}. 
Unfortunately, \importance{importance measures} that are claimed to have strong theoretical foundations and are widely used in practice \citep{Bhatt2020} often later turn out to be questionable \citep{Hooker2019,Kindermans2019,Adebayo2018,Jain2019,Wiegreffe2020}.

Accurately measuring if an explanation is faithful is therefore paramount. Such \emph{faithfulness} metrics are difficult to develop as the models are too complex to know what the correct explanation is. \citet{Doshi-Velez2017a} says a \emph{faithfulness} metric should use ``some formal definition of interpretability as a proxy for explanation quality.''

In this work, we use the definition of \emph{faithfulness} by \citet{Samek2017} and \citet{Hooker2019}: if information (input tokens) is truly important, then removing it should result in a worse model performance compared to removing random information (tokens).
We build upon the ROAR metric by \citet{Hooker2019}, which adds that it is necessary to retrain the model after information is removed, to avoid out-of-distribution issues. Finally, the model performance is compared with removing random information.

A limitation of ROAR is that it is theoretically impossible to measure the faithfulness of an \importance{importance measure} when dataset redundancies exist. For example, if two tokens are equally relevant but only one of them is identified as important, ROAR fails to remove the second token.

We propose \emph{Recursive ROAR} which solves this limitation. In addition to the \emph{Recursive ROAR} metric, we introduce a summarizing metric (RACU) which aggregates the results into a scalar metric. We hope that such a metric will make it more feasible to compare importance measures across papers.

Using the proposed faithfulness metrics, we perform a comprehensive comparative study of 4 different \importance{importance measures} and two popular architectures: BiLSTM-Attention and RoBERTa \citep{Liu2019}. We use 8 different datasets which are commonly used in the faithfulness of \emph{attention} literature \citep{Jain2019}. 

%To summarize we:
%\begin{itemize}[noitemsep]
%    \item Develop the faithfulness metric \emph{Recursive ROAR}.
%    \item Propose a summarizing scalar metric.
%    \item Use these metrics to perform a comprehensive comparative study of current \importance{importance measures}.
%\end{itemize}

Our comparative study reveals that no \importance{importance measure} is consistently better than others. 
Instead, we find that faithfulness is both task and model dependent.
This is valuable knowledge, as although each \importance{importance measure} might be equal in faithfulness, they are not equal in computational requirements or understandability to humans.

In particular, we find that \attention{attention} generally provides more sparse explanations than \gradient{gradient} or \ig{integrated gradient}. Although their faithfulness may be the same, a sparser explanation is often easier for humans to understand \citep{Miller2019}.

Computationally speaking, \ig{integrated gradient} is approximately 50 times more expensive than the \gradient{gradient} method. This additional complexity is usually justified by being considered more faithful than \gradient{gradient}. However, our results indicate that this is rarely a worthwhile trade-off.

\section{Related Work}
\label{sec:related-work}

Much recent work in NLP has been devoted to investigating the faithfulness of \importance{importance measures}, particularly \attention{attention}. In this section, we categorize these faithfulness metrics according to their underlying principle and discuss their drawbacks. ROAR \citep{Hooker2019} and our Recursive ROAR metrics differ significantly from these approaches.

The works on \attention{attention} are all based on the BiLSTM-Attention models and datasets from \citet{Jain2019}, they are therefore highly comparable. We use the same models and datasets, while also analyzing RoBERTa.

\subsection{Comparing with alternative importance measures}
The idea is to compare \attention{attention} with an alternative \importance{importance measure}, such as \gradient{gradient}. The claim is, if there is a correlation this would validate \attention{attention's} faithfulness. \citet{Jain2019} specifically compare with the \gradient{gradient} method and the \emph{leave-one-out} method. \citet{Meister2021a} repeat this experiment in a broader context.

Both \citet{Jain2019} and \citet{Meister2021a} find that there is little correlation between \importance{importance measures} and interpret this as attention being not faithful.

\citet{Jain2019} does acknowledge the limitations of this approach, as the alternative \importance{importance measures} are not themselves guaranteed to be faithful. A correlation, or lack of correlation, does therefore not inform about faithfulness. A criticism that we agree with and highlight here.

\subsection{Mutate attention to deceive}
\citet{Jain2019} propose that if there exist alternative attention weights that produce the same prediction, \attention{attention} is unfaithful.

They implement this idea by directly mutating the attention such that there is no prediction change but a large change in \attention{attention} and find that alternative attention distributions exist. \citet{Vashishth2019} and \citet{Meister2021a} apply a similar method and achieve similar results.

\citet{Wiegreffe2020} find this analysis problematic because the attention distribution is changed directly, thereby creating an out-of-distribution issue. This means that the new attention distribution may be impossible to obtain naturally from just changing the input, and it therefore says little about the faithfulness of attention.

\subsection{Optimize model to deceive}
Because the \emph{mutate attention to deceive} approach has been criticized for using direct mutation, an alternative idea is to learn an adversarial \attention{attention}.

\citet{Wiegreffe2020} investigate maximizing the KL-divergence between normal attention and adversarial attention while minimizing the prediction difference between the two models. By varying the allowed prediction difference, they show that it is not possible to significantly change the attention weights without affecting performance. Importantly, \citet{Wiegreffe2020} only use this experiment to invalidate the \emph{mutate attention to deceive} experiments, not to measure faithfulness. However, \citep{Meister2021a} do use this experiment setup as a faithfulness metric.

\citet{Pruthi2020} perform a similar analysis but report a contradictory finding. They find it is possible to significantly change the attention weights without affecting performance. They use this to show that attention is not faithful.

We find this approach problematic because by changing the optimization criteria the analysis is no longer about the standard BiLSTM-attention model \citep{Jain2019}, which is the subject of interest. Therefore, this analysis only works as a criticism of the \emph{mutate attention to deceive} approach, not as an evaluation of faithfulness.

\subsection{Known explanations in synthetic tasks}

\citet{Arras2020} constructs a purely synthetic task, where the true explanation is known. Evaluating importance measures against this true explanation serves as the faithfulness metric. Unfortunately, this approach cannot be used on real datasets and assumes a well behaved model.

\citet{Bastings2021}, a concurrent work to ours, therefore introduce spurious correlations into real datasets, creating partially synthetic tasks. They then evaluate if importance measures can detect these correlations. They conclude, similar to us, that faithfulness is both model and task-dependent.

We believe that this approach is the most valid among the mentioned metrics in the section. However, model behavior, and thereby the explanation behavior, can be drastically different on observations with spurious correlations from those without. This method is therefore limited in scope as it can only evaluate if the importance measure can be used to detect known spurious correlations.

\section{ROAR: RemOve And Retrain}
\label{sec:roar}

To address the shortcomings of the current faithfulness measures as described in \Cref{sec:related-work}, we base our metric on ROAR \citep{Hooker2019}.

ROAR has been used in computer vision to evaluate the faithfulness of \importance{importance measures} and to a limited extent in NLP \citep{Pham2021a}. The central idea is that if information is truly important, then removing it from the dataset and retraining a model on this reduced dataset should worsen model performance. This can then be compared with an uninformative baseline, where information is removed randomly.

For example, at a step size of $10\%$, one can remove the top-$\{10\%, 20\%, \cdots 90\%\}$ allegedly important tokens, evaluate the model performance, and compare this with removing $\{10\%, 20\%, \cdots 90\%\}$ random tokens. If the \importance{importance measures} is faithful, the former should result in a worse model performance than the latter.

This section covers how ROAR is adapted to an NLP context. Furthermore, we explain the dataset redundancy issue which is solved by our proposed Recursive ROAR metric. Finally, we show that Recursive ROAR is an improvement on ROAR using a synthetic task.

\subsection{Adaptation to NLP}
ROAR was originally proposed as a faithfulness metric in computer vision. In this context, pixels measured to be important are ``removed'' by replacing them with an uninformative value, such as a gray pixel \citep{Hooker2019}.

In this work, ROAR is applied to sequence classification tasks. Because these models use tokens, the uninformative value is a special \texttt{[MASK]} token (example in \Cref{fig:roar-masking-example}). We choose a \texttt{[MASK]} token rather than removing the token to keep the sequence length, which is an information source unrelated to \importance{importance measures}. 

\begin{figure}[h]
    \centering
    \footnotesize
    \resizebox{\linewidth}{!}{\begin{tabular}{p{0.2cm}p{7.8cm}}
    \toprule
 0\% & %
 \colorbox{rgb,255:red,254; green,228; blue,191}{\strut The}\allowbreak%
 \colorbox{rgb,255:red,253; green,191; blue,136}{\strut movie}\allowbreak%
 \colorbox{rgb,255:red,255; green,247; blue,237}{\strut is}\allowbreak%
 \colorbox{rgb,255:red,252; green,141; blue,89}{\strut great}\allowbreak%
 \colorbox{rgb,255:red,254; green,239; blue,217}{\strut .}\allowbreak%
 \colorbox{rgb,255:red,255; green,247; blue,237}{\strut I}\allowbreak%
 \colorbox{rgb,255:red,254; green,239; blue,217}{\strut really}\allowbreak%
 \colorbox{rgb,255:red,255; green,247; blue,237}{\strut liked}\allowbreak%
 \colorbox{rgb,255:red,254; green,239; blue,217}{\strut it}\allowbreak%
 \colorbox{rgb,255:red,255; green,247; blue,237}{\strut .} \\
 10\% & %
 \colorbox{white}{\strut The}\allowbreak%
 \colorbox{white}{\strut movie}\allowbreak%
 \colorbox{white}{\strut is}\allowbreak%
 \colorbox{white}{\strut [MASK]}\allowbreak%
 \colorbox{white}{\strut .}\allowbreak%
 \colorbox{white}{\strut I}\allowbreak%
 \colorbox{white}{\strut really}\allowbreak%
 \colorbox{white}{\strut liked}\allowbreak%
 \colorbox{white}{\strut it}\allowbreak%
 \colorbox{white}{\strut .} \\
 20\% & %
 \colorbox{white}{\strut The}\allowbreak%
 \colorbox{white}{\strut [MASK]}\allowbreak%
 \colorbox{white}{\strut is}\allowbreak%
 \colorbox{white}{\strut [MASK]}\allowbreak%
 \colorbox{white}{\strut .}\allowbreak%
 \colorbox{white}{\strut I}\allowbreak%
 \colorbox{white}{\strut really}\allowbreak%
 \colorbox{white}{\strut liked}\allowbreak%
 \colorbox{white}{\strut it}\allowbreak%
 \colorbox{white}{\strut .} \\
    \bottomrule
\end{tabular}}
    \caption{Example of \textbf{ROAR}. The first sentence shows the importance of various tokens. The next two sentences demonstrate the proportion of important tokens replaced by \texttt{[MASK]}. Note, the second sentence is enough to infer the sentiment.}
    \label{fig:roar-masking-example}
\end{figure}

\subsection{Recursive ROAR}
\label{sec:roar:recursive-roar}

With ROAR there are two conclusions, either 1) the  \importance{importance measure} is to some degree faithful or, 2) the faithfulness is unknown. The former is observed when the model's performance is statistically significantly below the random baseline. In the latter case, \citet{Hooker2019} explain that the \importance{importance measure} can either be not faithful or there can be a dataset redundancy. Recursive ROAR solves this redundancy issue and thereby provides a more informative conclusion.

A dataset redundancy affects the conclusion because the model does not need to use the redundant information. A faithful importance measure would therefore not highlight redundancies as important. After the important information which the importance measure did highlight is removed and the model is retrained, the redundant information can still keep the model's performance high. An example of this issue is demonstrated in \Cref{fig:roar-masking-example}.

We solve this issue by recursively recomputing the \importance{importance measure} at each iteration of information removal. This way, if the \importance{importance measure} is faithful, it would quickly mark the redundant information as important after which it would be removed. Note that already masked tokens are kept masked. We call this Recursive ROAR and provide an example in \Cref{fig:roar-redundancy-example}.

\begin{figure}[t]
    \centering
    \footnotesize
    \resizebox{\linewidth}{!}{\begin{tabular}{p{0.2cm}p{7.8cm}}
    \toprule
 0\% & %
 \colorbox{rgb,255:red,254; green,228; blue,191}{\strut The}\allowbreak%
 \colorbox{rgb,255:red,253; green,191; blue,136}{\strut movie}\allowbreak%
 \colorbox{rgb,255:red,255; green,247; blue,237}{\strut is}\allowbreak%
 \colorbox{rgb,255:red,252; green,141; blue,89}{\strut great}\allowbreak%
 \colorbox{rgb,255:red,254; green,239; blue,217}{\strut .}\allowbreak%
 \colorbox{rgb,255:red,255; green,247; blue,237}{\strut I}\allowbreak%
 \colorbox{rgb,255:red,254; green,239; blue,217}{\strut really}\allowbreak%
 \colorbox{rgb,255:red,255; green,247; blue,237}{\strut liked}\allowbreak%
 \colorbox{rgb,255:red,254; green,239; blue,217}{\strut it}\allowbreak%
 \colorbox{rgb,255:red,255; green,247; blue,237}{\strut .} \\
 10\% & %
 \colorbox{rgb,255:red,254; green,239; blue,217}{\strut The}\allowbreak%
 \colorbox{rgb,255:red,254; green,228; blue,191}{\strut movie}\allowbreak%
 \colorbox{rgb,255:red,255; green,247; blue,237}{\strut is}\allowbreak%
 \colorbox{rgb,255:red,254; green,228; blue,191}{\strut [MASK]}\allowbreak%
 \colorbox{rgb,255:red,255; green,247; blue,237}{\strut .}\allowbreak%
 \colorbox{rgb,255:red,254; green,239; blue,217}{\strut I }\allowbreak%
 \colorbox{rgb,255:red,253; green,191; blue,136}{\strut really}\allowbreak%
 \colorbox{rgb,255:red,252; green,141; blue,89}{\strut liked}\allowbreak%
 \colorbox{rgb,255:red,254; green,228; blue,191}{\strut it}\allowbreak%
 \colorbox{rgb,255:red,255; green,247; blue,237}{\strut .} \\
 20\% & %
 \colorbox{rgb,255:red,254; green,239; blue,217}{\strut The}\allowbreak%
 \colorbox{rgb,255:red,255; green,191; blue,136}{\strut movie}\allowbreak%
 \colorbox{rgb,255:red,255; green,247; blue,237}{\strut is}\allowbreak%
 \colorbox{rgb,255:red,254; green,228; blue,191}{\strut [MASK]}\allowbreak%
 \colorbox{rgb,255:red,255; green,247; blue,237}{\strut .}\allowbreak%
 \colorbox{rgb,255:red,254; green,239; blue,217}{\strut I }\allowbreak%
 \colorbox{rgb,255:red,254; green,228; blue,191}{\strut really}\allowbreak%
 \colorbox{rgb,255:red,255; green,247; blue,237}{\strut [MASK]}\allowbreak%
 \colorbox{rgb,255:red,254; green,239; blue,217}{\strut it}\allowbreak%
 \colorbox{rgb,255:red,255; green,247; blue,237}{\strut .} \\
    \bottomrule
\end{tabular}}
    \caption{Example of how a redundancy can be removed in \textbf{Recursive ROAR} by reevaluating the \importance{importance measure}. Compare this to \Cref{fig:roar-masking-example}, where redundancies are not removed and the performance can remain the same, even when the \importance{importance measure} is faithful.}% Note that the redundant information may be expressed in more complex ways than demonstrated here.}
    \label{fig:roar-redundancy-example}
\end{figure}

Note, Recursive ROAR might not remove all redundancies unless the step size is one token. However, because ROAR requires retraining the model, for every evaluation step, this is infeasible. Instead, we approximate it by removing a relative number of tokens. We discuss this more in \Cref{sec:appendix:absolute-roar}.

\subsection{Validation on a synthetic problem}

To show that Recursive ROAR provides an optimal faithfulness metric, we validate it on the same generated synthetic problem (with input $\mathbf{x}$ and output $y$) presented in the original ROAR paper \citep{Hooker2019}:
\begin{equation}
    \mathbf{x} = \frac{\mathbf{a} z}{10} + \mathbf{d} \eta + \frac{\epsilon}{10}, \quad y = \begin{cases} 1 & z > 0 \\
    0 & z \le 0 \end{cases}.
    \label{eq:roar:synthetic}
\end{equation}

Quoting \citet{Hooker2019} ``All random variables were sampled from a standard normal distribution. The vectors $\mathbf{a}$ and $\mathbf{d}$ are 16 dimensional vectors that were sampled once to generate the dataset. In $\mathbf{a}$ only the first 4 values have nonzero values to ensure that there are exactly 4 informative features. The values $z$, $\eta$, and $\epsilon$ are sampled independently for each example.''

The ground truth removal order is to remove the first 4 features (the specific order does not matter) followed by the remaining irrelevant features. Note that these first 4 features are mutually redundant.

In \citet{Hooker2019}, they do not use a specific importance measure. Instead, they use predefined removal orders. This avoids the redundancy issue in the synthetic task, although they do mention it as a limitation. Instead, we use the weights of a linear model as the importance measure and apply ROAR and Recursive ROAR using this explanation.

\begin{figure}[t]
    \centering
    \includegraphics[width=\linewidth]{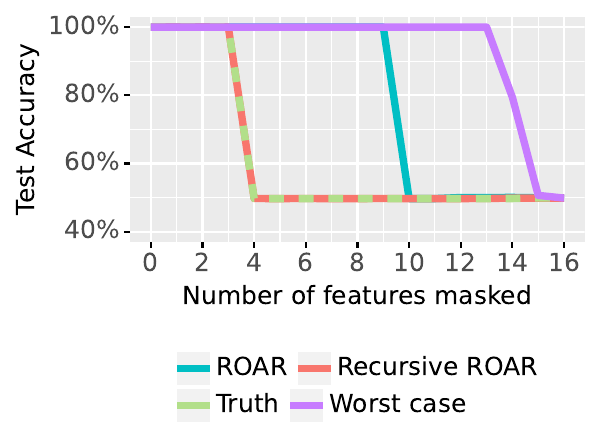}
    \caption{Using the weights of a linear model as the explanation, ROAR and Recursive ROAR are applied to the problem described in \eqref{eq:roar:synthetic}. In addition, the ground truth and worst case are shown. Recursive ROAR and the ground truth are identical.}
    \label{fig:roar:synthetic}
\end{figure}

\Cref{fig:roar:synthetic} shows that Recursive ROAR is identical to the ground truth, while ROAR is worse. %Meaning, ROAR is between the ground truth and the worst case.

\section{Importance Measures}
\label{sec:importance_measures}
In this section, we describe the importance measures that will be evaluated. We choose these explanations as they are common and computationally feasible to evaluate on every observation.

As \attention{attention} does not attend to the begin-of-sequence token, end-of-sequence token, and auxiliary sequence in paired-sequence problems, these tokens are also not considered for other importance measures. This is to ensure a fair comparison.

\paragraph{Attention} These are the attention weights of a BiLSTM-Attention model. We repeat the definitions in \Cref{sec:appendix:models:lstm}.

While we also look at a transformer-based model which also have internal attention mechanisms, these models do not provide one specific way to convert attention scores into an importance measure. There are proposals to turn the many attention heads into an importance measure \citep{Abnar2020a}. However, these are computationally expensive and requires knowing which layer to select. Performing this analysis is a standalone research topic which we will not answer. %It has also been suggested by \citet{Bastings2020} this direction is unproductive and efforts should be focused on gradient-based methods instead.

\paragraph{Gradient} Let the logits be denoted as $f(\mathbf{x})$. Then the gradient explanation is $\nabla_\mathbf{x} f(\mathbf{x})$, where $\mathbf{x}$ is a one-hot-encoding of the input \citep{Baehrens2010, Li2016}. To reduce away the vocabulary dimension, we use an $L_2$-norm.

\paragraph{Input times Gradient} This explanation is $\mathbf{x} \odot \nabla_\mathbf{x} f(\mathbf{x})$. Note that because $\mathbf{x}$ is a one-hot encoding, only one element per token will be non-zero. This non-zero element is considered as the explanation.

\paragraph{Integrated Gradient (IG)} \citet{Sundararajan2017a} argue this to be more faithful, via axiomatic proofs, compared to previous gradient-based methods. A disadvantage is that it is significantly more computationally intensive as it requires computing $k$ gradients. We use $k = 50$ like the original paper \citep{Sundararajan2017a}, and use $\mathbf{b} = \mathbf{0}$ as is done in NLP literature \citep{Mudrakarta2018}:
\begin{equation}
\begin{aligned}    
\operatorname{IG}(\mathbf{x}) &= (\mathbf{x} - \mathbf{b}) \odot \frac{1}{k} \sum_{i=1}^{k} \nabla_{\tilde{\mathbf{x}}_i} f(\tilde{\mathbf{x}}_i)_c \\
\tilde{\mathbf{x}}_i &= \mathbf{b} + \frac{i}{k}(\mathbf{x} - \mathbf{b}).
\end{aligned}
\end{equation}

\section{Experiments}
\label{sec:experiments}

The datasets, performance metrics, and the BiLSTM-attention model are identical to those used in \citet{Jain2019} and most other literature evaluating the faithfulness of \attention{attention}. In addition,  we use the RoBERTa-base model with the standard fine-tuning procedure \citep{Liu2019}. Details are in \Cref{sec:appendix:models}\makeatletter\ifacl@finalcopy\footnote{Code is available at \url{https://github.com/AndreasMadsen/nlp-roar-interpretability}}\else{\footnote{Code is available at \url{https://anonymous.4open.science/r/nlp-roar-interpretability-4802}. Will be made non-anonymized in camera ready version.}
}\fi\makeatother.

We report model performance on the 8 studied datasets in \Cref{tab:results:performance}. Below, we provide a short description of each dataset. We provide additional details in \Cref{sec:appendix:datasets}.
\begin{enumerate}[itemsep=0pt, nolistsep, noitemsep]
    \item Two sentiment tasks: SST \citep{Socher2013} and IMDB \citep{Maas2011}.
    \item Two tasks with long-sequences: Diabetes and Anemia \citep{Johnson2016}. These datasets contain many redundancies.
    \item A paired-sequence class: SNLI \citep{Bowman2015}. 
    \item \textit{bAbI} \citep{Weston2015} task 1 to 3. These are synthetic paired-sequence problems.
\end{enumerate}

\begin{table}[ht]
    \centering
    \begin{tabular}{lccc} 
\toprule   
Dataset & Sequence & \multicolumn{2}{c}{Performance [\%]} \\
\cmidrule(r){3-4}
& length & LSTM & RoBERTa \\
\midrule   
Anemia & 2267 & $88^{+1.1}_{-2.2}$ & $86^{+0.6}_{-0.7}$ \\
Diabetes & 2207 & $81^{+2.2}_{-2.9}$ & $76^{+0.7}_{-0.6}$ \\
IMDB & 181 & $90^{+0.4}_{-0.7}$ & $95^{+0.2}_{-0.2}$ \\
SNLI & 16 & $78^{+0.2}_{-0.3}$ & $91^{+0.1}_{-0.1}$ \\
SST & 20 & $82^{+0.6}_{-1.0}$ & $94^{+0.3}_{-0.3}$ \\
bAbI-1 & 38 & $100^{+0.0}_{-0.1}$ & $100^{+0.0}_{-0.0}$ \\
bAbI-2 & 96 &  $68^{+9.1}_{-19.1}$ & $100^{+0.1}_{-0.1}$ \\
bAbI-3 & 308 & $60^{+6.5}_{-4.9}$ & $81^{+6.8}_{-20.0}$ \\
\bottomrule
\end{tabular}

    \caption{Model performance scores and sequence-length for each dataset. Performance is averaged over 5 seeds with a 95\% confidence interval. Following \citet{Jain2019}, we report performance as macro-F1 for SST, IMDB, Anemia and Diabetes, micro-F1 for SNLI, and accuracy for bAbI.\\
    The average sequence-length is for the BiLSTM-attention model, for the RoBERTa model the number will be higher but with inputs truncated at 512 tokens.}
    \label{tab:results:performance}
\end{table}

\subsection{Supporting experiments}
In \Cref{sec:appendix:classical-roar}, we compare \emph{ROAR} and \emph{Recursive ROAR}. These results show dataset redundancies interfere with \emph{ROAR}. For example, with the Diabetes dataset, only by using \emph{Recursive ROAR} can \emph{gradient} be measured to be faithful.

In \Cref{sec:appendix:absolute-roar}, we avoid the approximation of removing a relative number of tokens at 10\% increments by instead removing exactly one token in each iteration. These results show that the approximation does affect the results, but not the conclusions that can be drawn from the results.

In \Cref{sec:appendix:sparsity}, we report the sparsity of each importance measure and find that \attention{attention} is significantly more sparse than other importance measures. If the faithfulness is equal, this may make it more desirable as sparse explanations are more understandable to humans \citep{Miller2019}.

\subsection{Main experiment: Recursive ROAR}
\label{sec:results:roar}

\begin{figure*}[tb!]
    \centering
    \includegraphics[trim=0.1in 0.6cm 0.1in 0, clip, width=\linewidth]{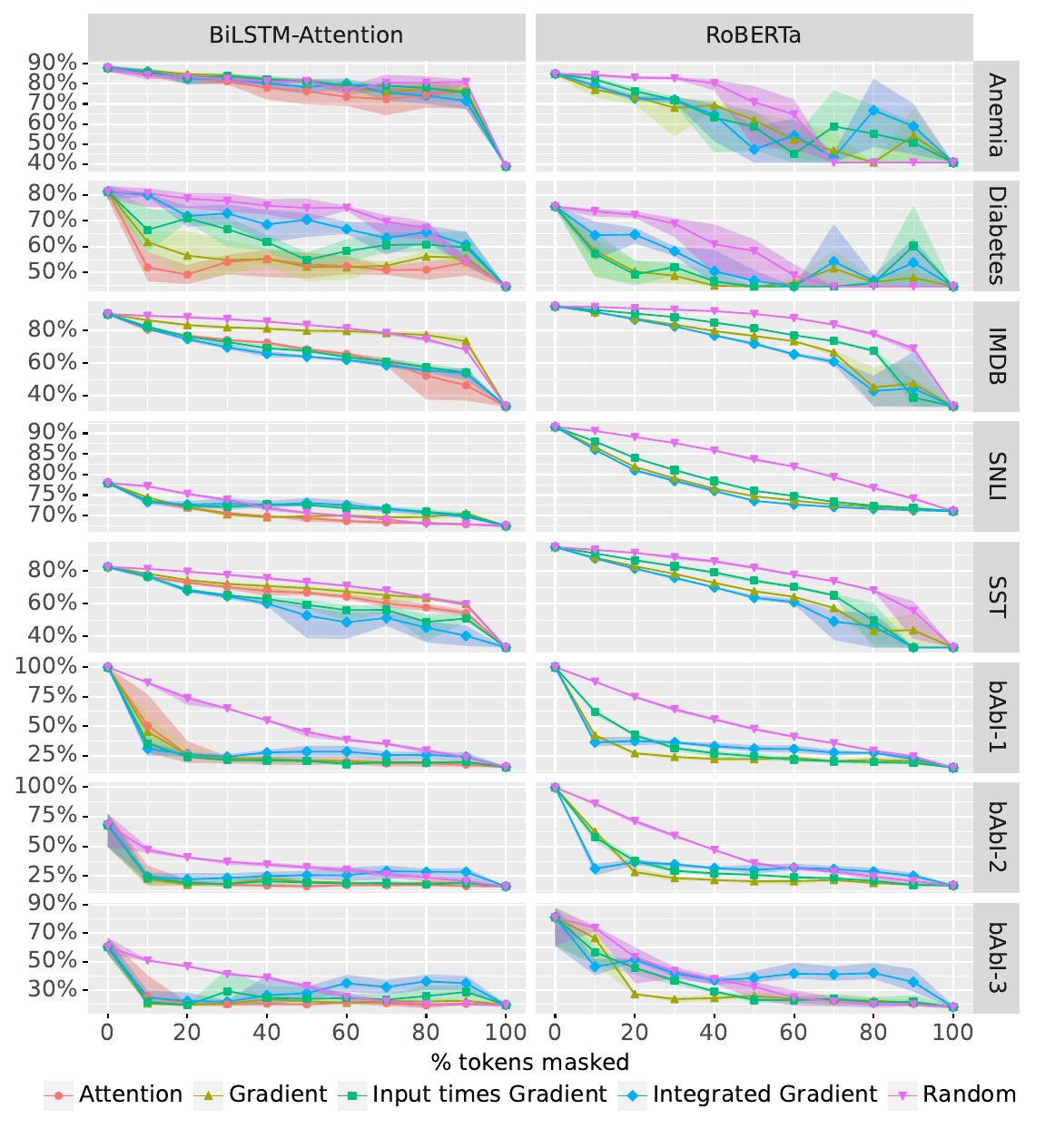}
    \caption{Recursive ROAR results, showing model performance at x\% of tokens masked. A model performance below \emph{random} indicates faithfulness, while above or similar to \emph{random} indicates a non-faithful importance measure. Performance is averaged over 5 seeds with a 95\% confidence interval.}
    \label{fig:roar}
\end{figure*}

To evaluate the faithfulness of importance measures, we apply \emph{Recursive ROAR} to all datasets and both models. The results are presented in \Cref{fig:roar} and discussed in \Cref{sec:findings}.

In \Cref{sec:appendix:compute}, we report the compute times. Because BiLSTM-Attention is a small model and RoBERTa-base is only fine-tuned, Recursive ROAR is feasible when \importance{importance measure} can be evaluated on every observation. For some \importance{importance measures}, like SHAP \citep{Lundberg2017}, which have exponential compute complexity, ROAR would not be feasible. Additionally, for large language models, like T5 \citep{Raffel2020}, ROAR would also be difficult to apply as fine-tuning these models is generally challenging.

\subsubsection{How to interpret}

If the model performance of a given \importance{importance measure} is below the random baseline, then this indicates a faithful importance measure. Note that ``faithful'' is not absolute, rather we measure the degree of faithfulness. However, if the model performance is not statistical significant below the random baseline, then the \importance{importance measure} is not considered to be faithful. With the \emph{(Not Recursive) ROAR} measure, this latter case would be inconclusive as the faithfulness could be hidden by dataset redundancies.

\Cref{fig:roar} also presents the model performance at 100\% masking, which provides a lower bound for the model performance and is helpful as the datasets are often biased. These biases come from unbalanced classes or the secondary sequence for the paired-sequence tasks \citep{Gururangan2018}. For these datasets, sequence-length bias is not a concern \Cref{sec:appendix:datasets:biases}.

\subsection{Summarizing faithfulness metric}
While a ROAR plot can provide valuable insights, such as ``this importance measure is only faithful for the top-20\% most important tokens,'' it does not summarize the faithfulness to a scalar metric. Such a metric is useful as it allows for easy comparisons, particularly between different papers. 

\begin{figure}[h!]
    \centering
    \includegraphics[width=\linewidth]{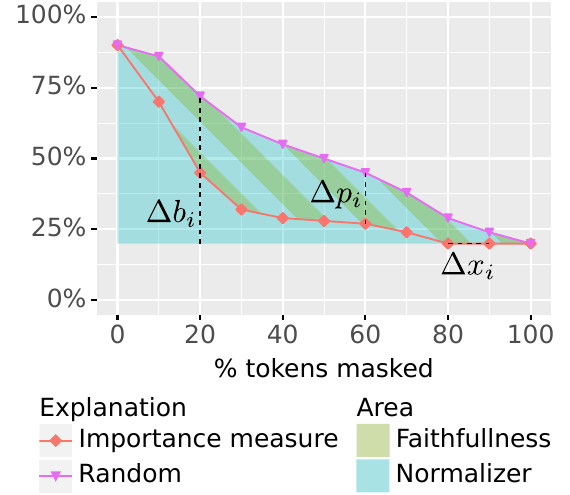}
    \caption{Visualization of the faithfulness calculation done in \eqref{eq:faithfulness-metric}. The \emph{faithfulness} area is the numerator in \eqref{eq:faithfulness-metric}, while the \emph{normalizer} area is the denominator. Essentially \eqref{eq:faithfulness-metric} computes the \textbf{r}elative \textbf{a}rea-between-\textbf{cu}rves (RACU) between an \emph{explanation} curve and the \emph{random} baseline curve.}
    \label{fig:faithfulness-metric-visualization}
\end{figure}

To provide a scalar metric, we propose using a \textbf{r}elative \textbf{a}rea-between-\textbf{cu}rves (RACU) metric. Intuitively, an importance measure is more faithful if it has a larger area between the random baseline curve and the importance measure curve. Additionally, when the importance measure is above the random baseline, a negative area is contributed. Finally, the metric is normalized by an upper bound, where the performance at 100\% masking is achieved immediately. A visualization of this calculation can be seen in \Cref{fig:faithfulness-metric-visualization}.

Using an area-between-curves is useful because, unlike many other summarizing statistics, it is invariant to the step-size used in ROAR. In this case, we have a step size of $10\%$. Future work may choose a smaller or larger step size depending on their computational resources.

Let $r_i$ be the masking ratio at step $i$ out of $I$ total step, in our case $r = \{0\%, 10\%, \cdots, 100\%\}$. Let $p_i$ be the model performance for a given importance measure and $b_i$ be the random baseline performance. With this, the metric is defined in \eqref{eq:faithfulness-metric}, and we present the results in \Cref{tab:faithfulness-metric}.

\begin{equation}
\begin{aligned}
    \operatorname{RACU} &= \frac{
        \sum_{i=1}^{I-1} \frac{1}{2} \Delta x_i (\Delta p_i + \Delta p_{i+1})
    }{
        \sum_{i=1}^{I-1} \frac{1}{2} \Delta x_i (\Delta b_i + \Delta b_{i+1})
    } \\
    \text{where } \Delta x_i &= x_{i+1} - x_i \quad \textit{step size} \\
    \Delta p_i &= b_i - p_i \quad \textit{performance delta} \\
    \Delta b_i &= b_i - b_I \quad \textit{baseline delta} 
\end{aligned}
\label{eq:faithfulness-metric}
\end{equation}

\begin{table}[ht!]
\centering
\resizebox{\linewidth}{!}{\begin{tabular}{llcc}
\toprule
        & Importance & \multicolumn{2}{c}{RACU Faithfulness [\%]} \\
\cmidrule(lr){3-4}
Dataset & Measure & LSTM & RoBERTa \\
\midrule   
\multirow[c]{4}{*}{Anemia} & Attention & $7.6^{+7.9}_{-6.8}$ & -- \\
 & Gradient & $1.0^{+2.8}_{-4.1}$ & $18.2^{+11.8}_{-13.8}$ \\
 & $\mathbf{x}\ \odot \text{Gradient}$ & $0.8^{+2.5}_{-3.5}$ & $8.8^{+22.7}_{-22.8}$ \\
 & IG & $4.9^{+2.7}_{-1.8}$ & $12.5^{+11.3}_{-7.0}$ \\
\cmidrule{1-4}
\multirow[c]{4}{*}{Diabetes} & Attention & $66.5^{+6.5}_{-13.0}$ & -- \\
 & Gradient & $57.4^{+7.8}_{-7.0}$ & $57.9^{+14.4}_{-19.8}$ \\
 & $\mathbf{x}\ \odot \text{Gradient}$ & $33.7^{+7.0}_{-15.7}$ & $53.4^{+23.2}_{-29.3}$ \\
 & IG & $11.4^{+8.4}_{-15.0}$ & $26.1^{+12.0}_{-25.1}$ \\
\cmidrule{1-4}
\multirow[c]{4}{*}{IMDB} & Attention & $29.8^{+5.0}_{-3.4}$ & -- \\
 & Gradient & $3.1^{+2.4}_{-3.3}$ & $25.4^{+3.1}_{-2.0}$ \\
 & $\mathbf{x}\ \odot \text{Gradient}$ & $28.4^{+1.0}_{-0.9}$ & $16.9^{+1.1}_{-3.0}$ \\
 & IG & $32.5^{+0.9}_{-1.0}$ & $35.1^{+2.4}_{-1.7}$ \\
\cmidrule{1-4}
\multirow[c]{4}{*}{SNLI} & Attention & $36.5^{+3.0}_{-3.5}$ & -- \\
 & Gradient & $18.7^{+5.1}_{-3.5}$ & $50.7^{+1.1}_{-0.8}$ \\
 & $\mathbf{x}\ \odot \text{Gradient}$ & $-10.7^{+6.1}_{-5.7}$ & $41.0^{+0.4}_{-0.5}$ \\
 & IG & $-13.9^{+5.0}_{-5.0}$ & $56.7^{+1.0}_{-1.1}$ \\
\cmidrule{1-4}
\multirow[c]{4}{*}{SST} & Attention & $15.7^{+2.4}_{-2.4}$ & -- \\
 & Gradient & $7.6^{+2.3}_{-2.0}$ & $26.1^{+1.6}_{-2.2}$ \\
 & $\mathbf{x}\ \odot \text{Gradient}$ & $28.0^{+5.6}_{-4.4}$ & $18.6^{+4.1}_{-4.6}$ \\
 & IG & $37.8^{+4.6}_{-5.3}$ & $32.9^{+1.8}_{-1.5}$ \\
\cmidrule{1-4}
\multirow[c]{4}{*}{bAbI-1} & Attention & $66.5^{+9.2}_{-9.2}$ & -- \\
 & Gradient & $66.1^{+5.9}_{-6.5}$ & $64.2^{+2.6}_{-2.6}$ \\
 & $\mathbf{x}\ \odot \text{Gradient}$ & $71.2^{+4.0}_{-4.2}$ & $52.1^{+1.8}_{-3.7}$ \\
 & IG & $59.1^{+6.8}_{-7.4}$ & $48.2^{+4.1}_{-5.7}$ \\
\cmidrule{1-4}
\multirow[c]{4}{*}{bAbI-2} & Attention & $75.4^{+4.9}_{-8.1}$ & -- \\
 & Gradient & $66.3^{+4.2}_{-5.1}$ & $57.8^{+2.0}_{-2.0}$ \\
 & $\mathbf{x}\ \odot \text{Gradient}$ & $66.7^{+8.0}_{-12.4}$ & $48.1^{+3.2}_{-3.5}$ \\
 & IG & $34.6^{+13.4}_{-14.8}$ & $42.0^{+3.8}_{-4.8}$ \\
\cmidrule{1-4}
\multirow[c]{4}{*}{bAbI-3} & Attention & $77.7^{+9.6}_{-8.1}$ & -- \\
 & Gradient & $73.0^{+9.1}_{-7.6}$ & $34.0^{+14.6}_{-15.1}$ \\
 & $\mathbf{x}\ \odot \text{Gradient}$ & $53.9^{+10.7}_{-24.1}$ & $22.4^{+15.9}_{-12.4}$ \\
 & IG & $25.9^{+8.5}_{-9.1}$ & $-27.9^{+18.0}_{-49.1}$ \\
\bottomrule
\end{tabular}
}
\caption{Faithfulness metric defined as a \textbf{r}elative \textbf{a}rea-between-\textbf{cu}rves (RACU) using Recursive ROAR, see \eqref{eq:faithfulness-metric}. Higher values mean more faithful, zero or negative values mean distinctly not faithful. IG is an acronym for \emph{Integrated Gradient}. $\mathbf{x} \odot \text{Gradient}$ refers to \emph{Input times Gradient}.}
\label{tab:faithfulness-metric}
\end{table}

\section{Important Findings}
\label{sec:findings}

Based on the results in \Cref{fig:roar} and \Cref{tab:faithfulness-metric}, we highlight the following important findings.

\paragraph{Faithfulness is model-dependent.}
In particular, the faithfulness with SNLI is highly model-dependent as seen in \Cref{tab:faithfulness-metric}. Furthermore, comparing the faithfulness between the two models,the faithfulness of \emph{Gradient} on IMDB and \emph{Integrated Gradient} on bAbI-3 is significantly affected by the model architecture.

\paragraph{Faithfulness is task-dependent.}
For BiLSTM-Attention, in \Cref{tab:faithfulness-metric}, \emph{Attention} is best for SNLI while \emph{Input times Gradient} and \emph{Integrated Gradient} is best for SST.

For RoBERTa, \emph{Integrated Gradient} is best for IMDB and SNLI, while \emph{Gradient} is best for bAbI-1 and bABI-2. In fact, \emph{Integrated Gradient} is worst in all bAbI tasks.

%LSTM-Anemia: Attention and Integrated Gradient (barely).
%LSTM-Diabetes: Attention and Gradient
%LSTM-IMDB: Attention and IG
%*LSTM-SNLI: Attention
%*LSTM-SST: Input times gradient and IG
%LSTM-bAbI-2 and 3 Integrated gradient is worst

%*RoBERTa-IMDB: Integrated Gradient
%*RoBERTa-SNLI: Integrated Gradient
%RoBERTa-SST: Gradient and Integrated Gradient
%*RoBERTa-bAbI-1 and 2: Gradient

\paragraph{Attention can be faithful.}
In \Cref{tab:faithfulness-metric}, \emph{Attention} is among the top explanations in terms of faithfulness, except for SST. This contradicts many of the previous results mentioned in \Cref{sec:related-work}, which found attention to be unfaithful.

Because attention is computationally free and attention is more sparse (\Cref{sec:appendix:sparsity}), which is important for human understanding \citep{Miller2019}, attention can be an attractive explanation.

\paragraph{Integrated Gradient is not necessarily more faithful than Gradient or Input times Gradient.}
For BiLSTM-Attention, in \Cref{tab:faithfulness-metric}, bAbI-2, bAbI-3, and SNLI has least one gradient-based importance measure which is significantly more faithful than \emph{Integrated Gradient}. For RoBERTa, we find the same for bAbI-2 and bAbI-3.
These results contradicts the claim that Integrated Gradient is theoretically superior \citep{Sundararajan2017a}. This is a valuable finding, as Integrated Gradient is significantly more computationally expensive than other gradient-based importance measure.

\paragraph{Importance measures often work best for the top-20\% most important tokens.}
In \Cref{fig:roar}, we observe that the largest drop tends to happen at about 10\% or 20\% tokens masked. This indicates that importance measures are best at ranking the most important tokens, while for less important tokens, they become noisy. This is particularly observed in bAbI for both models and Diabetes with the BiLSTM-Attention model.

\paragraph{Class leakage can cause the model performance to increase.}
Because the importance measures explain predictions of the target label, they can leak the target label when allegedly important tokens are masked.

Consider a sentiment classification task. If an importance measure indicates that the word \emph{bad} is a strong indicator of negative sentiment, then in the next iteration \emph{bad} would be masked in negative sentences. This means the presence of \emph{bad} now leaks the true label (positive sentiment) which may increase the performance.

This issue is particularly observed with bAbI-3 using RoBERTa in \Cref{fig:roar}, where the performance increases slightly at 60\% tokens masked. This issue affects both ROAR and Recursive ROAR (\Cref{sec:appendix:classical-roar}). In fact, it likely affects most faithfulness metrics. However, Recursive ROAR can mitigate this issue to some extent. We discuss this more in \Cref{sec:appendix:leakage}.

\makeatletter
\ifacl@finalcopy
\fi
\makeatother
\section{Conclusion}

We show that Recursive ROAR is an improvement on ROAR. In a synthetic setting, Recursive ROAR matches the ground truth, while ROAR does not. Additionally, we argue why other faithfulness metrics may be either invalid or limited in scope.

We then use Recursive ROAR to measure the faithfulness of the most common importance measures, including attention. This is done on both recurrent and transformer-based neural models.
% We provide a list of the most important findings in \Cref{sec:findings}. 
In general, we find that the faithfulness of importance measures is both model-dependent and task-dependent. This means that no general recommendation can be made for NLP practitioners considering the current importance measures. Instead, it is necessary to measure the faithfulness of different importance measures given a task and a model.

Because Recursive ROAR works on real-world datasets and not just synthetic problems, we hope it can serve as a standardized benchmark for the faithfulness of importance measures in NLP.

\section{Limitations}
\label{sec:limitations}

Recursive ROAR requires the model to be retrained. This means it is not possible to evaluate the faithfulness of a specific model instance, rather we evaluate the faithfulness of the model architecture. The confidence intervals we provide then inform us about what can be statistically expected in terms of the faithfulness for a model instance.

The retraining dependence also means Recursive ROAR can only measure the faithfulness of a task-model combination that is feasible to train/fine-tune repeatedly and importance measures that are feasible to compute across the entire dataset.

A second category of limitation comes from the use of masking. In particular, if the dataset is heavily biased, then the performance at 100\% will remain high. This can happen if for example the sequence length is a good predictor of the class. In principle, this means that no tokens are important. Therefore, we can't comment on the faithfulness of an importance measure in that context. In such a case, the faithfulness metric in \eqref{eq:faithfulness-metric} should become unstable (in theory division by zero, but in practice chaotic values) and result in a large confidence interval.

As discussed in the previous section, because the importance measures explains the target class, they can leak the class information when used to mask input features. This can make an importance measure appear less faithful than it actually is. However, this issue cannot make an importance measure appear more faithful than it is (see \Cref{sec:appendix:leakage} for more discussion).

Furthermore, while we believe Recursive ROAR provides a useful metric for faithfulness, only measuring faithfulness is not enough for an explanation to be used in production settings \citep{Doshi-Velez2017a}. In addition to faithfulness, one should also evaluate if the explanation is understandable to humans (known as human-groundedness). This is already being done to some extent but is a complex topic \citep{Sen2020a,Hase2020a,Prasad2021,Gonzalez2021,Schuff2022,Lertvittayakumjorn2019,Nguyen2018}.

Finally, \citet{Doshi-Velez2017a} argue that explanations should be tested with the final application in mind. Unfortunately, in deployment settings very little evaluation of any kind is done \citep{Bhatt2020}. However, we hope that this work can help establish a metric for faithfulness.

\section*{Impact Statement and Ethics}

Interpretability itself is paramount to the ethical deployment of machine learning models. Whether this is to proactively ensure that a model performs predictions that align with human values or to retroactively understand what went wrong in a model's prediction \citep{Doshi-Velez2017a, Doshi-Velez2017}.

Providing misleading explanations can be potentially dangerous, as even wrong explanations can be very convincing. To prevent this we need accurate faithfulness metrics, which this paper hopes to provide. However, history has shown that it is notoriously difficult to develop principled faithfulness metrics \citep{Jain2019, Kindermans2019, Adebayo2018, Hooker2019}.

It is always a possibility that a proposed faithfulness metric is flawed, including the one proposed here. If this is not caught it could lead to more misleading explanations. To prevent this, we try to be extra transparent about the limitations of the proposed faithfulness metric, as described in \Cref{sec:limitations}. In particular, we also advocate for testing an interpretability method in terms of the human-groundedness and application-groundedness before using it in production \citep{Doshi-Velez2017a}.

\makeatletter
\ifacl@finalcopy
\section*{Acknowledgements}
SR is supported by the Canada CIFAR AI Chairs program and the NSERC Discovery Grant program. Computing resources were provided by Compute Canada.
\fi
\makeatother

\bibliography{references}

%\clearpage
\appendix
\section{Explanation of class leakage}
\label{sec:appendix:leakage}

When importance measures are computed, it is the prediction of the gold label that is explained. For example, for the \emph{Gradient} method, it is $\nabla_\mathbf{x} f(x)_y$ that is computed, where $\mathbf{x}$ is the input and $y$ is the gold label.

We want an importance measure for the correct label, as removing the tokens that are relevant for making a wrong prediction, would help the performance of the model. If the gold label was not used, the faithfulness results would be affected by the model performance. As faithfulness and model performance should be unrelated, this is not a desired outcome.

This is a general issue with faithfulness metrics due to how importance measures are calculated in benchmark settings. This is an unfortunate gap between the benchmark-setting and the practical setting where the gold label is unknown. Furthermore, it is rarely documented.

In ROAR and Recursive ROAR, this issue is expressed as an increase in the model performance. Intuitively, it should not be possible for the model performance to increase with more information removed compared to less. However, because the importance measures are w.r.t. the gold label, they can leak the gold label which can increase the model performance.

\paragraph{Thought experiment.} Consider the SST dataset, a binary sentiment classification task. Let's say that the \texttt{and} token has a spurious correlation with the positive label (there is some truth to this). Although, clearly the \texttt{and} token can appear in both negative and positive sentences.

For example, let's say that just using the \texttt{and} token provides a 60\% accurate classification of positive labels. An importance measure would therefore highlight the \texttt{and} token as being important for the prediction of positive sentiment. Unfortunately, an importance measure might not consider the \texttt{and} token equally important for a negative sentiment (could be due to non-linearity). If all \texttt{and} tokens are removed from sentences with positive sentiment as the gold label, the existence of an \texttt{and} token is now a perfect predictor of negative sentiment. Hence, the model performance will increase (there will still be negative sentiment sentences without \texttt{and} tokens).

Assuming a faithful importance measure, in the next iteration of Recursive ROAR the \texttt{and} token would now be important for predicting negative sentiment and would be removed. However, this assumption is rarely completely justified, there is also no guarantee that \texttt{and} is considered the most important for all observations. Finally, in the case where a relative number of tokens are masked, the removal of other tokens may leak the gold label.

\paragraph{General issue.} As mentioned, the need to use the gold label is a general issue that likely\footnote{We could not find any documentation for which label is used in relevant non-ROAR metrics, and no code has been published.} extends beyond ROAR. However, because ROAR presents a more qualitative metric (\Cref{fig:roar}) where a curve can be observed to increase, this issue is more apparent. Had we just presented the summarizing metric (\Cref{tab:faithfulness-metric}), as most faithfulness metrics do, the issue would have been hidden.

\section{Datasets}
\label{sec:appendix:datasets}

\begin{table*}[tb!]
\centering
\begin{tabular}{lccccccc} 
\toprule   
Dataset & \multicolumn{3}{c}{Size} & \multicolumn{3}{c}{Performance [\%]} \\
\cmidrule(r){2-4} \cmidrule(r){5-7}
        & Train & Validation & Test & LSTM by \citet{Jain2019} & LSTM & RoBERTa \\
\midrule
Anemia & 4262 & 729 & 1242 & $92$ & $88^{+1.1}_{-2.2}$ & $86^{+0.6}_{-0.7}$ \\
Diabetes & 8066 & 1573 & 1729 & $79$ & $81^{+2.2}_{-2.9}$ & $76^{+0.7}_{-0.6}$ \\
IMDB & 17212 & 4304 & 4362 & $78$ & $90^{+0.4}_{-0.7}$ & $95^{+0.2}_{-0.2}$ \\
SNLI & 549367 & 9842 & 9824 & $88$ & $78^{+0.2}_{-0.3}$ & $91^{+0.1}_{-0.1}$ \\
SST & 6579 & 848 & 1776 & $81$ & $82^{+0.6}_{-1.0}$ & $94^{+0.3}_{-0.3}$ \\
bAbI-1 & 8500 & 1500 & 1000 & $100$ & $100^{+0.0}_{-0.1}$ & $100^{+0.0}_{-0.0}$ \\
bAbI-2 & 8500 & 1500 & 1000 & $48$ & $68^{+9.1}_{-19.1}$ & $100^{+0.1}_{-0.1}$ \\
bAbI-3 & 8500 & 1500 & 1000 & $62$ & $60^{+6.5}_{-4.9}$ & $81^{+6.8}_{-20.0}$ \\
\bottomrule
\end{tabular}
\caption{Datasets statistics for single-sequence and paired-sequence tasks. Following \citet{Jain2019}, we use the same BiLSTM-attention model and report performance as macro-F1 for SST, IMDB, Anemia and Diabetes, micro-F1 for SNLI, and accuracy for bAbI.}
\label{tab:appendix-dataset:details}
\end{table*}

The datasets used in this work are listed below. All datasets are public works. There have been made no attempts to identify any individuals. The use is consistent with their intended use and all tasks were already established by \citet{Jain2019}.

The MIMIC-III dataset \citep{Johnson2016} is an anonymized dataset of health records. To access this a HIPAA certification is required, which the first author has obtained. Additionally, the MIMIC-III data has not been shared with anyone else, including other authors of this paper. 

Below, we provide more details on each dataset. In \Cref{tab:appendix-dataset:details}, we provide dataset statistics.

\subsection{Single-sequence tasks}
\begin{enumerate}[noitemsep]
\item \textit{Stanford Sentiment Treebank (SST)} \citep{Socher2013} -- Sentences are classified as positive or negative. The original dataset has 5 classes. Following \citet{Jain2019}, we label (1,2) as negative, (4,5) as positive, and ignore the neural sentences.

\item \textit{IMDB Movie Reviews} \citep{Maas2011} -- Movie reviews are classified as positive or negative.

\item \textit{MIMIC (Diabetes)} \citep{Johnson2016} -- Uses health records to detect if a patient has Diabetes.

\item \textit{MIMIC (Chronic vs Acute Anemia)} \citep{Johnson2016} -- Uses health records to detect whether a patient has chronic or acute anemia. 
\end{enumerate}

\subsection{Paired-sequence tasks}
\begin{enumerate}[resume, noitemsep]
\item \textit{Stanford Natural Language Inference (SNLI)} \citep{Bowman2015} -- Inputs are premise and hypothesis. The hypothesis either entails, contradicts, or is neutral w.r.t. the premise.

\item \textit{bAbI} \citep{Weston2015} -- A set of artificial text for understanding and reasoning. We use the first three tasks, which consist of questions answerable using one, two, and three sentences from a passage, respectively.
\end{enumerate}

\subsection{Class bias and sequence-length bias}
\label{sec:appendix:datasets:biases}

Because Recursive ROAR masks tokens the sequence-length remains the same. At 100\% masking the only information the model has is the sequence-length. To understand the relevance of the sequence-length, we compare the 100\% masking model performance with a basic class-majority classifier. The results in \Cref{tab:apppendix-dataset:sequence-length} show that the sequence-length does not have much relevance. SNLI does show significant difference but this relates it's the secondary sequence being a very good predictor on its own, not the sequence length \citep{Gururangan2018}. 

\begin{table}[h]
    \centering
    \begin{tabular}{lccc}
\toprule
Dataset & Majority & LSTM & RoBERTa \\
\midrule
Anemia & $39\%$ & $39\%^{+0.0\%}_{-0.0\%}$ & $41\%^{+0.0\%}_{-0.0\%}$ \\
Diabetes & $45\%$ & $45\%^{+0.0\%}_{-0.0\%}$ & $45\%^{+0.0\%}_{-0.0\%}$ \\
IMDB & $34\%$ & $33\%^{+0.1\%}_{-0.4\%}$ & $33\%^{+0.1\%}_{-0.3\%}$ \\
SNLI & $34\%$ & $67\%^{+0.3\%}_{-0.3\%}$ & $71\%^{+0.1\%}_{-0.1\%}$ \\
SST & $33\%$ & $33\%^{+0.0\%}_{-0.0\%}$ & $33\%^{+0.0\%}_{-0.0\%}$ \\
bAbI-1 & $15\%$ & $15\%^{+0.8\%}_{-0.6\%}$ & $15\%^{+0.0\%}_{-0.0\%}$ \\
bAbI-2 & $19\%$ & $16\%^{+0.3\%}_{-0.4\%}$ & $17\%^{+0.4\%}_{-0.4\%}$ \\
bAbI-3 & $19\%$ & $20\%^{+0.8\%}_{-1.1\%}$ & $18\%^{+1.2\%}_{-0.9\%}$ \\
\bottomrule
\end{tabular}

    \caption{Performance of the class-majority classifier and the BiLSTM-Attention and RoBERTa classifier on the 100\% masked dataset. Performance is the standard metric for the dataset. Meaning, macro-F1 for SST, IMDB, Anemia and Diabetes, micro-F1 for SNLI, and accuracy for bAbI.}
    \label{tab:apppendix-dataset:sequence-length}
\end{table}

\section{Models}
\label{sec:appendix:models}

\subsection{BiLSTM-Attention}
\label{sec:appendix:models:lstm}
The BiLSTM-Attention models, hyperparameters, and pre-trained word embeddings are the same as those from \citet{Jain2019}. We repeat the configuration details in \Cref{tab:appendix-models:lstm-details}. 

There are two types of models, single-sequence and paired-sequence, however, they are nearly identical. They only differ in how the context vector $\mathbf{b}$ is computed.
In general, we refer to $\mathbf{x} \in \mathbb{R}^{T \times V}$ as the one-hot encoding of the primary input sequence, of length $T$ and vocabulary size $V$. The logits are then $f(\mathbf{x})$ and the target class is denoted as $c$.

\subsubsection{Single-sequence}
A $d$-dimentional word embedding followed by a bidirectional LSTM (BiLSTM) encoder is used to transform the one-hot encoding into the hidden states $\mathbf{h}_x \in \mathbb{R}^{T \times 2 d}$. These hidden states are then aggregated using an additive attention layer $\mathbf{h}_\alpha = \sum_{i=1}^{T}\alpha_i \mathbf{h}_{x,i}$.

To compute the attention weights $\alpha_i$ for each token:
\begin{equation}
    \alpha_i = \frac{\text{exp}(\mathbf{u}^\top_i \mathbf{v})}{\sum_j\text{exp}(\mathbf{u}^\top_j \mathbf{v})}, \
    u_i = \text{tanh}(\mathbf{W} \mathbf{h}_{x,i} + \mathbf{b})
\end{equation}
where $\mathbf{W},\mathbf{b},\mathbf{v}$ are model parameters. Finally, the $\mathbf{h}_\alpha$ is passed through a fully-connected layer to obtain the logits $f(\mathbf{x})$.

\subsubsection{Paired-sequence}
For paired-sequence problems, the two sequences are denoted as $\mathbf{x} \in \mathbb{R}^{T_x \times V}$ and $\mathbf{y} \in \mathbb{R}^{T_y \times V}$. The inputs are then transformed to embeddings using the same embedding matrix, and then transformed using two separate BiLSTM encoders to get the hidden states, $\mathbf{h}_x$ and $\mathbf{h}_y$. Likewise, they are aggregated using additive attention $\mathbf{h}_\alpha = \sum_{i=1}^{T_x}\alpha_i \mathbf{h}_{x,i}$.

The attention weights $\alpha_i$ are computed as:
\begin{equation}
\begin{aligned}
    \alpha_i &= \frac{\text{exp}(\mathbf{u}^\top_i \mathbf{v})}{\sum_j\text{exp}(\mathbf{u}^\top_j \mathbf{v})} \\
    \mathbf{u}_i &= \text{tanh}(\mathbf{W}_x \mathbf{h}_{x,i} + \mathbf{W}_y \mathbf{h}_{y, T_2}),
\end{aligned}
\end{equation}

where $\mathbf{W}_x, \mathbf{W}_y, \mathbf{v}$ are model parameters. Finally, $\mathbf{h}_\alpha$ is transformed with a dense layer.

\begin{table*}[tb!]
\centering
\resizebox{\textwidth}{!}{\begin{tabular}{lcp{3cm}cccc} 
\toprule   
Dataset & Variant & Embedding initialization & Embedding size & nb. of parameters & Batch size & Max epochs \\
\midrule
Anemia & Singe & Word2Vec trained on MIMIC & 300 & 5 352 158 & 32 & 8 \\
Diabetes & Single & Word2Vec trained on MIMIC & 300 & 6 138 158 & 32 & 8 \\
IMDB & Single & Pretrained FastText & 300 & 4 218 458 & 32 & 8 \\
SNLI & Paired & Pretrained Glove (840B) & 300 & 13 601 939 & 128 & 25 \\
SST & Single & Pretrained FastText & 300 & 4 603 658 & 32 & 8 \\
bAbI-1 & Paired & Standard Normal Distribution & 50 & 55 048 & 50 & 100 \\
bAbI-2 & Paired & Standard Normal Distribution & 50 & 55 048 & 50 & 100 \\
bAbI-3 & Paired & Standard Normal Distribution & 50 & 55 048 & 50 & 100 \\
\bottomrule
\end{tabular}
}
\caption{Details on the BiLSTM-attention models' hyperparameters. Everything is exactly as done by \citet{Jain2019}. For all datasets, ASMGrad Adam \citep{Reddi2018} is used with default hyperparameters ($\lambda=0.001$, $\beta_1=0.9$, $\beta_2=0.999$, $\epsilon=10^{-8}$) and a weight decay of $10^{-5}$.}
\label{tab:appendix-models:lstm-details}
\end{table*}

\subsection{RoBERTa}

We use RoBERTa \citep{Liu2019} as a transformer-based model due to its consistent convergence. Consistent convergence is helpful as ROAR and Recursive ROAR requires the model to be trained many times. We use the RoBERTa-base pre-trained model and only perform fine-tuning. The hyperparameters are those defined used by \citet[Appendix C]{Liu2019} on GLUE tasks. We list the hyperparameters in \Cref{tab:appendix-models:roberta-details}.

\begin{table}[h!]
\centering
\begin{tabular}{lcc} 
\toprule   
Dataset & Variant & Max epochs \\
\midrule
Anemia & Single & 3 \\
Diabetes & Single & 3 \\
IMDB & Single & 3 \\
SNLI & Paired & 3 \\
SST & Single & 3 \\
bAbI-1 & Paired & 8 \\
bAbI-2 & Paired & 8 \\
bAbI-3 & Paired & 8 \\
\bottomrule
\end{tabular}

\caption{Details on the RoBERTa models' hyperparameters. RoBERTa \citep{Liu2019} is fine-tuned using the RoBERTa-base pre-trained model from HuggingFace \citep{Wolf2019} (125M parameters). The hyperparameters are those used by \citet{Liu2019} on GLUE tasks \citep[Appendix C]{Liu2019}. The optimizer is AdamW \citep{Loshchilov2019}, the learning rate has linear decay with a warmup ratio of 0.06, and there is a weight decay of $0.01$. Additionally, we use a batch size of $16$ and a learning rate of $2\cdot 10^{-5}$.}
\label{tab:appendix-models:roberta-details}
\end{table}

RoBERTa makes use of a beginning-of-sequence $\texttt{[CLS]}$ token, a end-of-sequence $\texttt{[EOS]}$ token, a separation token $\texttt{[SEP]}$ token, and a masking token $\texttt{[MASK]}$ token. The masking token used during pre-training is the same token that we use for masking allegedly important tokens.

For the single-sequence tasks, we encode as \texttt{[CLS]} \textit{\dots sentence \dots} \texttt{[EOS]}. For the paired-sequence tasks, we encode as \texttt{[CLS]} \textit{\dots main sentence \dots} \texttt{[SEP]} \textit{\dots auxiliary sentence \dots} \texttt{[EOS]}. Note that when computing the importance measures, only the main sentence is considered. This is to be consistent with the BiLSTM-attention model.

\section{Compute}
\label{sec:appendix:compute}

In this section, we document the compute times and resources used for computing the results. Unfortunately, our compute infrastructure changed during the making of this paper. The BiLSTM-attention results were computed on V100 GPUs while the RoBERTa results were computed on A100 GPUs. The A100 GPU is significantly faster than the V100 GPU, hence the compute times are not comparable across models. We could have recomputed the BiLSTM-attention results, but doing so would be a waste of resources. We report the machine specifications in \Cref{tab:appendix-compute:spec}.

\begin{table}[h]
    \centering
    \begin{tabular}{lp{5cm}}
        \toprule
        & BiLSTM-attention \\
        \cmidrule{2-2}
        CPU & 4 cores, Intel Gold 6148 Skylake @ 2.4 GHz \\
        GPU & 1x NVidia V100 SXM2 (16 GB) \\
        Memory & 24 GB \\
        \midrule
        & RoBERTa \\
        \cmidrule{2-2}
        CPU & 6 cores, AMD Milan 7413 @ 2.65 GHz 128M cache L3 \\
        GPU & 1x NVidia A100 (40 GB) \\
        Memory & 24 GB \\
        \bottomrule
    \end{tabular}
    \caption{Compute hardware used for each model. Note, the models were computed on a shared user system. Hence, we only report the resources allocated for our jobs.}
    \label{tab:appendix-compute:spec}
\end{table}

The compute times are reported in \Cref{tab:appendix-compute:walltime}. All compute was done using 99\% hydroelectric energy.

While the totals in \Cref{tab:appendix-compute:walltime} may be large, in partial situations only one dataset is usually considered. Additionally, the variance in \Cref{fig:roar} is quite low, making less seeds an option. Finally, the compute time of \emph{integrated gradient} is approximately 2/3 of the total. As discussed in \Cref{sec:findings}, this is rarely worth it. Practical settings may want to not consider \emph{integrated gradient} at all for this reason.

\begin{table}[H]
\centering
\resizebox{\linewidth}{!}{\begin{tabular}{llcc}
\toprule
        & Importance & \multicolumn{2}{c}{Walltime [hh:mm]} \\
\cmidrule(r){3-4}
Dataset & Measure & LSTM & RoBERTa \\
\midrule   
\multirow[c]{5}{*}{Anemia} & Random & 00:09 & 00:03  \\
 & Attention & 00:09 & -- \\
 & Gradient & 00:11 & 00:04 \\
 & Input times Gradient & 00:11 & 00:04 \\
 & Integrated Gradient & 00:44 & 00:27 \\
\cmidrule{1-4}
\multirow[c]{5}{*}{Diabetes} & Random & 00:17 & 00:05  \\
 & Attention & 00:17 & -- \\
 & Gradient & 00:23 & 00:07 \\
 & Input times Gradient & 00:23 & 00:07 \\
 & Integrated Gradient & 01:46 & 01:09 \\
\cmidrule{1-4}
\multirow[c]{5}{*}{IMDB} & Random & 00:05 & 00:08 \\
 & Attention & 00:05 & -- \\
 & Gradient & 00:05 & 00:10 \\
 & Input times Gradient & 00:05 & 00:10 \\
 & Integrated Gradient & 00:20 & 02:10 \\
\cmidrule{1-4}
\multirow[c]{5}{*}{SNLI} & Random & 00:49 & 01:03 \\
 & Attention & 00:46 & -- \\
 & Gradient & 00:48 & 01:28 \\
 & Input times Gradient & 00:48 & 01:10 \\
 & Integrated Gradient & 01:09 & 05:41 \\
\cmidrule{1-4}
\multirow[c]{5}{*}{SST} & Random & 00:02 & 00:02 \\
 & Attention & 00:02 & -- \\
 & Gradient & 00:02 & 00:02 \\
 & Input times Gradient & 00:02 & 00:02 \\
 & Integrated Gradient & 00:03 & 00:06 \\
\cmidrule{1-4}
\multirow[c]{5}{*}{bAbI-1} & Random & 00:08 & 00:04 \\
 & Attention & 00:09 & -- \\
 & Gradient & 00:08 & 00:04 \\
 & Input times Gradient & 00:08 & 00:04 \\
 & Integrated Gradient & 00:10 & 00:11 \\
\cmidrule{1-4}
\multirow[c]{5}{*}{bAbI-2} & Random & 00:12 & 00:06 \\
 & Attention & 00:12 & -- \\
 & Gradient & 00:12 & 00:06 \\
 & Input times Gradient & 00:12 & 00:06 \\
 & Integrated Gradient & 00:15 & 00:32 \\
\cmidrule{1-4}
\multirow[c]{5}{*}{bAbI-3} & Random & 00:24 & 00:11 \\
 & Attention & 00:25 & -- \\
 & Gradient & 00:25 & 00:13 \\
 & Input times Gradient & 00:25 & 00:13 \\
 & Integrated Gradient & 00:32 & 01:12 \\
\midrule
\midrule
\multirow[c]{3}{*}{\textbf{Total}} & sum & 13:38 & 17:20 \\
& x9 iterations (approx.) & 5 days & 6.5 days \\
& x5 seeds (approx.) & 25.5 days & 32.5 days \\
\bottomrule
\end{tabular}
}
\caption{Compute times for each model and importance measure combination. Note, there is no need to compute models for each importance measure at 0\% and 100\% masking. Hence, we report for 9 iterations.}
\label{tab:appendix-compute:walltime}
\end{table}

\section{Sparsity}
\label{sec:appendix:sparsity}

In this section, we analyse the sparsity of each importance measure. While none of the importance measures produce an actual importance for any token, they may have most of the importance assigned to just a few tokens.

This analysis serves two purposes, to show that masking a relative number of tokens is justified and to test if any importance measure are more sparse than others.

\begin{figure*}[b!]
    \centering
    \includegraphics[trim=0.1in 0.6cm 0.1in 0, clip, width=\linewidth]{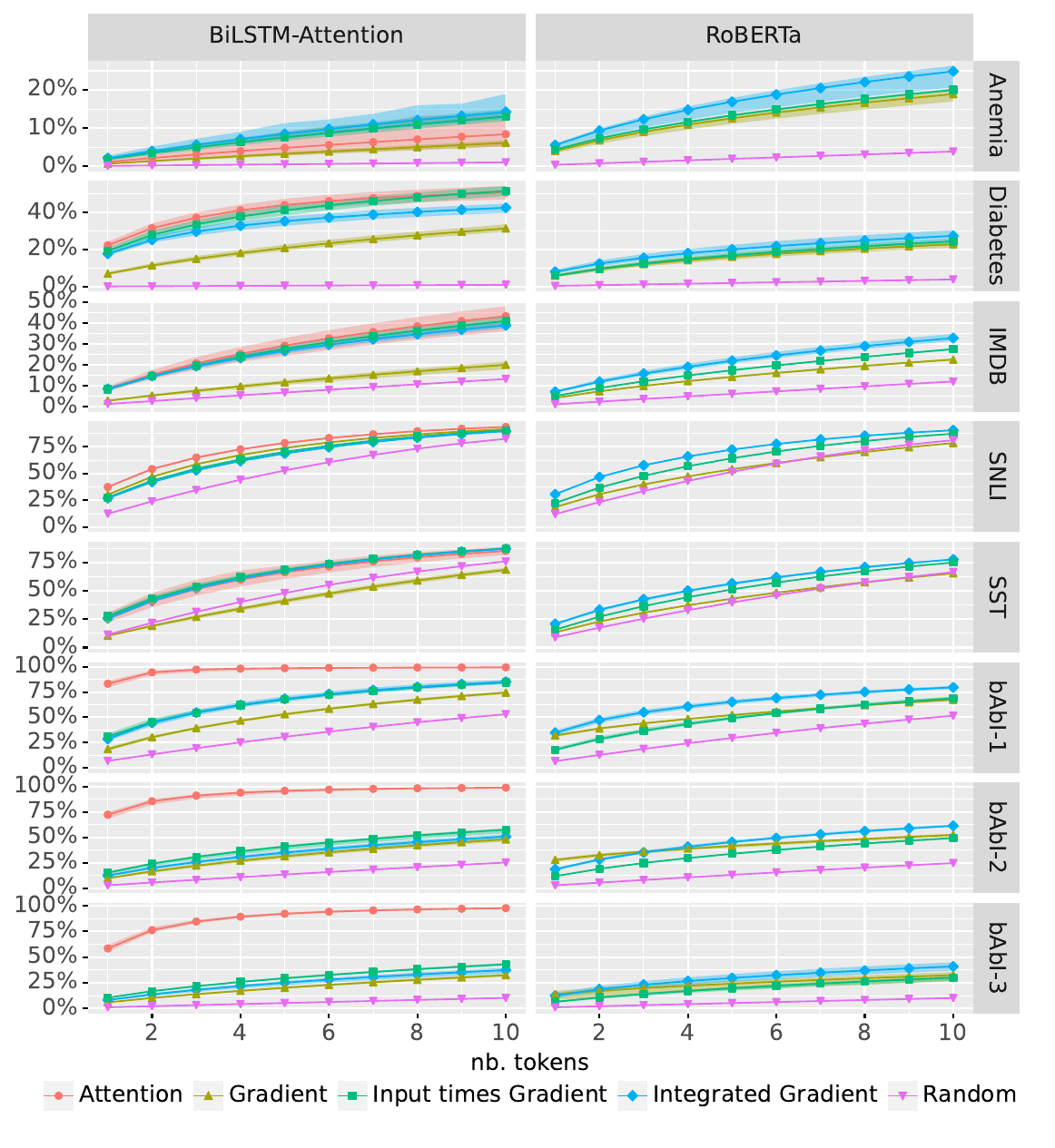}
    \caption{Shows the accumulative importance score relative to the total importance score, for the top-k number of tokens. The metric is averaged over 5 seeds with a 95\% confidence interval. Note that datasets are not equal in sequence-length, the scores are therefore hard to compare across datasets. Please refer to \Cref{tab:results:performance} for statistics on the sequence-length.}
    \label{fig:appendix:sparsity:absolute}
\end{figure*}

\paragraph{Masking a relative number of tokens is justified.}
If the majority of the importance is assigned to just a few tokens (e.g. 10 tokens have 99\% of the total importance scores), then it would make more sense to perform the non-approximate version of Recursive ROAR where exactly one token is masked in each iteration.

In \Cref{fig:appendix:sparsity:absolute}, we look at the sparsity considering the top-10 tokens. We find that that the sparsity is not sufficiently high to justify masking exactly one token in each iteration. For completeness, we include this analysis in \Cref{sec:appendix:absolute-roar}.

There are cases where masking exactly one token in each iteration could make sense, for example, for \emph{attention} in bAbI. However, as this is a comparative study among several importance measures and datasets, this is not enough. 

\begin{figure*}[tb!]
    \centering
    \includegraphics[trim=0.1in 0.6cm 0.1in 0, clip, width=\linewidth]{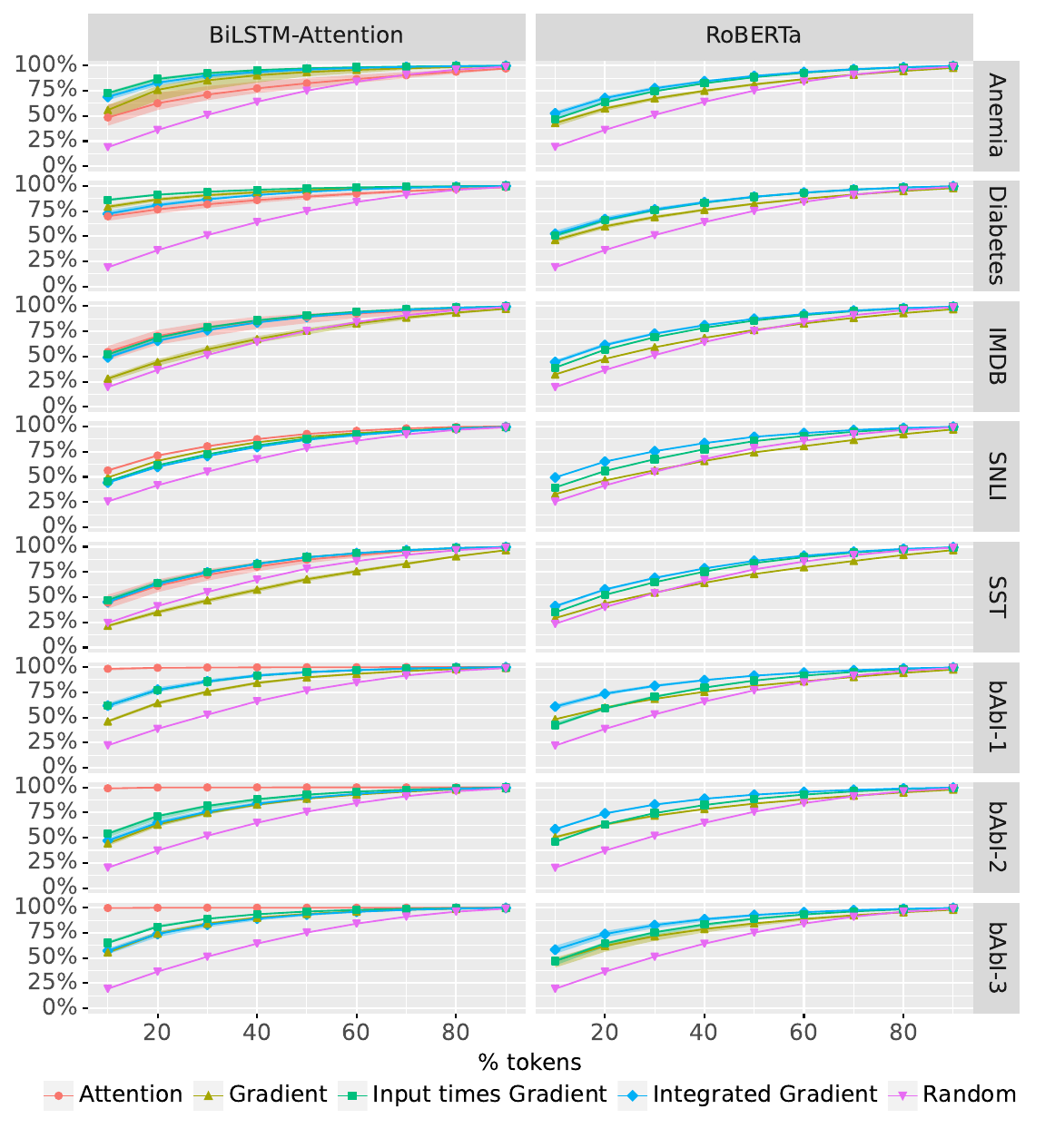}
    \caption{The accumulative importance score relative to the total importance score for the top-x\% number of tokens. The metric is averaged over 5 seeds with a 95\% confidence interval.}
    \label{fig:appendix:sparsity:relative}
\end{figure*}

\paragraph{Attention is more sparse than others importance measures}
If a particular importance measure is more sparse than others, while having a similar faithfulness, then the more sparse importance measure would be preferable. This is because it is more likely to be understandable to humans \citep{Miller2019}.

In \Cref{fig:appendix:sparsity:relative}, we look at the sparsity considering a relative number of tokens. We find that for some datasets, in particular bAbI, attention is the most sparse importance measure. Besides this, integrated gradient is usually the most sparse is nearly all cases. However, while the difference in sparsity is often statistically significant we speculate that the difference is not large enough to cause a difference in practical settings.

\section{Recursive ROAR with a stepsize of one token}
\label{sec:appendix:absolute-roar}

\begin{figure*}[!tb]
    \centering
    \includegraphics[trim=0.1in 0.6cm 0.1in 0, clip, width=\linewidth]{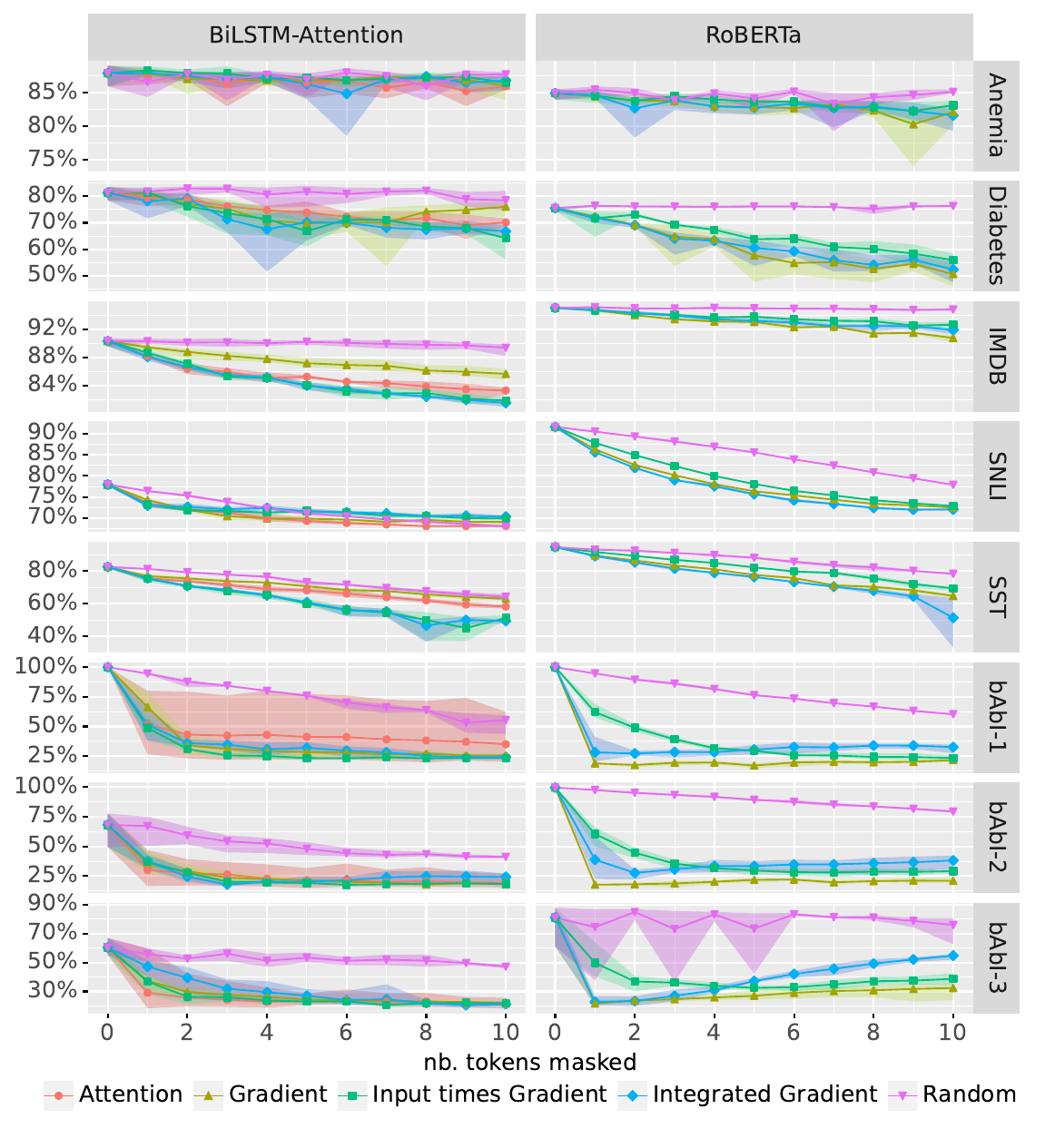}
    \caption{Recursive ROAR results, showing model performance at up to 10 tokens masked. Note that because the datasets have more than 10-tokens, the conclusion one can draw from this plot may change if more tokens were considered. However, in general, a model performance below \emph{random} indicates faithfulness, while above or similar to \emph{random} indicates a non-faithful importance measure. Performance is averaged over 5 seeds with a 95\% confidence interval.}
    \label{fig:appendix:absolute-roar}
\end{figure*}

To analyze the effect of masking 10\%, as opposed to masking exactly one token in each iteration, we perform the Recursive ROAR experiment with exactly one token token masked. The results are in \Cref{fig:appendix:absolute-roar}. Because this is computationally expensive, we only do this for up to 10 tokens. This makes it harder to make draw clear conclusions from this experiment, in particular because not all redundancies are removed when only masking 10 tokens.

In general, the results in \Cref{fig:appendix:absolute-roar} show that the approximation of masking 10\% in each iteration does affect the results. However, we can draw the same conclusions. That being said, some of the conclusions are less obvious because we only look at 10 tokens.

\subsection{The results are affected by the approximation}
Looking just at RoBERTa, for Diabetes, \emph{Integrated Gradient} yields 65\% performance at 10\% masking (approximately 51 tokens), while \emph{Integrated Gradient} yields 55\% performance at 10 tokens. Similarly for bAbI-3, \emph{Gradient} yields 65\% at 10\% masking (approximately 30 tokens), while \emph{Gradient} yields 30\% at 10 tokens. Both of these cases, shows that a lower performance is achieved earlier when masking one token in each iteration.

This is to be expected, as masking one token in each iteration is more effective for removing redundancies. Were we to complete the experiment to eventually mask all tokens, the faithfulness scores can therefore be expected to be higher.

\subsection{The conclusions are the same}

In \Cref{sec:findings}, we present 5 findings. Here, we briefly show that the same conclusions can be drawn from \Cref{fig:appendix:absolute-roar}. However, as only 10 tokens are masked they may be less obvious and there may be less evidence.

\paragraph{Faithfulness is model-dependent.} Yes, this is most clearly seen for IMDB, where BiLSTM-Attention archives significantly lower performance (higher faithfulness) compared to RoBERTa.
    
\paragraph{Faithfulness is task-dependent.} Yes, looking at BiLSTM-Attention, for IMDB \emph{Integrated Gradient} is the worst importance measure. However, for the bAbI tasks  \emph{Integrated Gradient} is among the best importance measures.
    
\paragraph{Attention can be faithful.} Yes, particularly for bAbI, IMDB, and Diabetes attention is faithful.
    
\paragraph{Integrated Gradient is not necessarily more faithful than Gradient or Input times Gradient.} Yes, considering BiLSTM-Attention, IMDB \emph{Integrated Gradient} is significantly worse than other explanations. For most datasets, \emph{Integrated Gradient} has similar faithfulness as other importance measures.
    
\paragraph{Importance measures often work best for the top-20\% most important tokens.} As \Cref{fig:appendix:absolute-roar} only shows 10 tokens, which is usually below top-20\% this is hard to comment on.
    
\paragraph{Class leakage can cause the model performance to increase.} For RoBERTa, in bAbI-3, the \emph{Integrated Gradient} importance measure can be seen to increase performance after 2 tokens are masked.

\section{ROAR vs Recursive ROAR}
\label{sec:appendix:classical-roar}

\begin{figure*}[tb!]
    \centering
    \includegraphics[trim=0.1in 0.6cm 0.1in 0, clip, width=\linewidth]{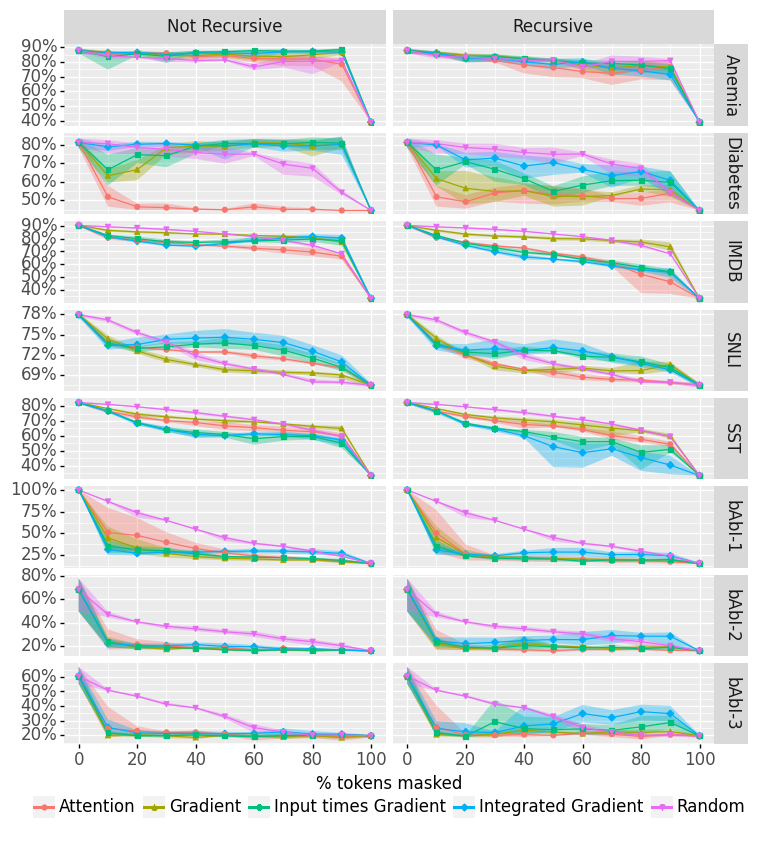}
    \caption{ROAR and Recursive ROAR results for \textbf{BiLSTM-Attention}, showing model performance at x\% of tokens masked. A model performance below \emph{random} indicates faithfulness. For Recursive ROAR a curve above or similar to \emph{random} indicates a non-faithful importance measure, while for ROAR by \citet{Hooker2019} this case is inconclusive. Performance is averaged over 5 seeds with a 95\% confidence interval.}
    \label{fig:appendix:classical-roar:rnn}
\end{figure*}

\begin{figure*}[tb!]
    \centering
    \includegraphics[trim=0.1in 0.6cm 0.1in 0, clip, width=\linewidth]{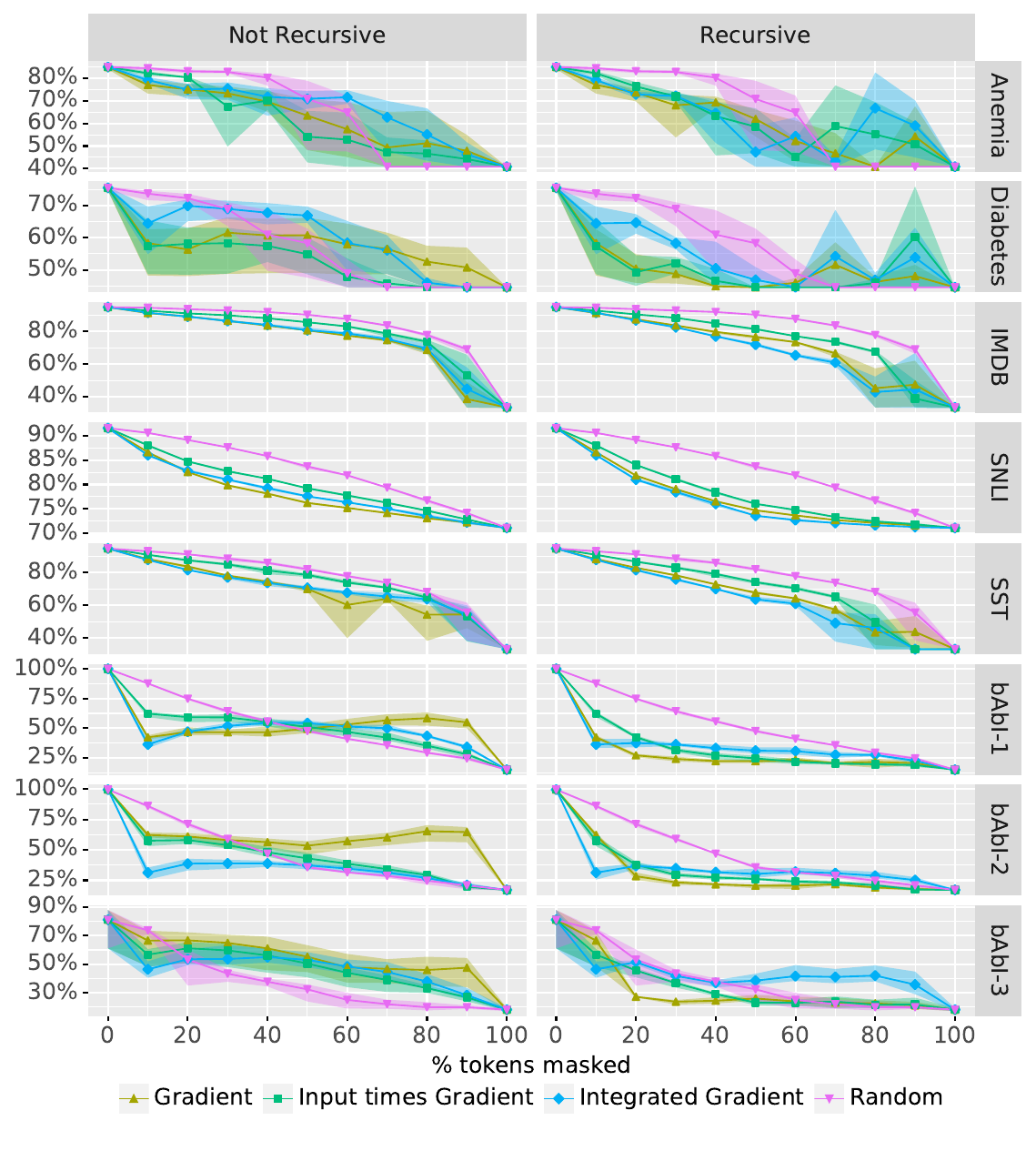}
    \caption{ROAR and Recursive ROAR results for \textbf{RoBERTa}, showing model performance at x\% of tokens masked. A model performance below \emph{random} indicates faithfulness. For Recursive ROAR a curve above or similar to \emph{random} indicates a non-faithful importance measure, while for ROAR by \citet{Hooker2019} this case is inconclusive. Performance is averaged over 5 seeds with a 95\% confidence interval.}
    \label{fig:appendix:classical-roar:roberta}
\end{figure*}

As an ablation study we compare ROAR by \citet{Hooker2019} with our Recursive ROAR. \Cref{fig:appendix:classical-roar:rnn} shows the comparison for BiLSTM-Attention and \Cref{fig:appendix:classical-roar:roberta} shows the comparison for RoBERTa. Recall that for ROAR by \citet{Hooker2019} it is not possible to say that an importance measure is not faithful. 

\paragraph{Some datasets have redundancies which affects ROAR.}
In particular, we find that Diabetes shows a significant difference comparing ROAR with Recursive ROAR. This is both for BiLSTM-Attention (\Cref{fig:appendix:classical-roar:rnn}) and RoBERTa (\Cref{fig:appendix:classical-roar:roberta}). For both models, \emph{Gradient} and \emph{Input times Gradient} becomes faithful with Recursive ROAR. Additionally, for RoBERTa the same is the case for \emph{Integrated Gradient}. This is not surprising, as Diabetes contains incredibly long sequences and contains redundancies.

Also, for IMDB, and to a lesser extent SST, there is a clear difference between BiLSTM-Attention and RoBERTa. This too is not surprising, as sentiment can often be inferred from just a single word. However, there are likely to be many positive or negative words in each observation.

%RNN BENIFIT:
%Diabetes, Gradient, input times gradient, becomes faithful. Also difference. Not surprising, as this datasets have many redundancies. 
%IMDB (SST) difference on all but Gradient. Again, sententiment tasks there may be a few relevant words. And only one of them is enougth.

%RoBERTa BENIFIT:
%Diabetese, Integrated, Gradient, input-times-gradient. Less than RNN.
%IMDB (SST) difference on all
%bAbI has huge difference on Gradient.

\paragraph{Class leakage affects both ROAR and Recursive ROAR.}
We observe the class leakage issue for ROAR in SNLI with BiLSTM-Attention and for the bAbI tasks with RoBERTa. We observe the issue for Recursive ROAR in bAbI with BiLSTM-Attention. The fact that the issue mostly exists with bAbI is somewhat encouraging, as the bAbI datasets are synthetic. The class leakage issue appears to affect real datasets less.

%RNN ISSUE:
%Both: SNLI, Integrated gradient
%Not Recursive. Gradient and Input times gradient.
%Recursive: bAbI (synthetic, maybe not a big deal).

%RoBERTA ISSUE:
%Issue is less pronounced, 

\end{document}